\documentclass[letterpaper]{article} 
\usepackage[]{aaai2026}  
\usepackage{times}  
\usepackage{helvet}  
\usepackage{courier}  
\usepackage[hyphens]{url}  
\usepackage{graphicx} 
\usepackage{natbib}  
\usepackage{caption} 
\usepackage{algorithm}
\usepackage{algorithmic}
\usepackage{booktabs}   
\usepackage{array} 

\usepackage{amsmath}
\usepackage{array}
\usepackage[table]{xcolor}

\usepackage{newfloat}
\usepackage{listings}
\DeclareCaptionStyle{ruled}{labelfont=normalfont,labelsep=colon,strut=off} 
\floatstyle{ruled}
\newfloat{listing}{tb}{lst}{}
\floatname{listing}{Listing}
\nocopyright 

\title{MoSE: Skill-by-Skill Mixture-of-Experts Learning for Embodied Autonomous Machines}

\author{
LU XU$^{1}$, JIAQIAN YU$^{1}$, XIONGFENG PENG$^{1}$, YIWEI CHEN$^{1}$, WEIMING LI$^{1}$,\\
JAEWOOK YOO$^{2}$, SUNGHYUN CHUNAG$^{2}$, DONGWOOK LEE$^{2}$, DAEHYUN JI$^{2}$, CHAO ZHANG$^{1}$\\
$^{1}$Advanced Research Lab, Samsung Research China-Beijing\\
$^{2}$Manufacturing/Material Handling AI Lab (DS AI Center)
}

\usepackage{bibentry}

\begin{document}

\maketitle

\begin{abstract}
To meet the growing demand for smarter, faster, and more efficient embodied AI solutions, we introduce a novel Mixture-of-Expert (MoE) method that significantly boosts reasoning and learning efficiency for embodied autonomous systems. General MoE models demand extensive training data and complex optimization, which limits their applicability in embodied AI such as autonomous driving (AD) and robotic manipulation. In this work, we propose a skill-oriented MoE called MoSE, which mimics the human learning and reasoning process skill-by-skill, step-by-step. We introduce a skill-oriented routing mechanism that begins with defining and annotating specific skills, enabling experts to identify the necessary competencies for various scenarios and reasoning tasks, thereby facilitating skill-by-skill learning. To better align with multi-step planning in human reasoning and in end-to-end driving models, we build a hierarchical skill dataset and pretrain the router to encourage the model to think step-by-step. Unlike other multi-round dialogues, MoSE integrates valuable auxiliary tasks (e.g. perception-prediction-planning for AD, and high-level and low-level planning for robots) in one single forward process without introducing any extra computational cost. With less than 3B sparsely activated parameters, our model effectively grows more diverse expertise and outperforms models on both AD corner-case reasoning tasks and robot reasoning tasks with less than 40\% of the parameters.

\end{abstract}


\section{Introduction}

\begin{figure}[h]
\includegraphics[trim={0 0.15in 0 0.1in},clip,width=0.5\textwidth]{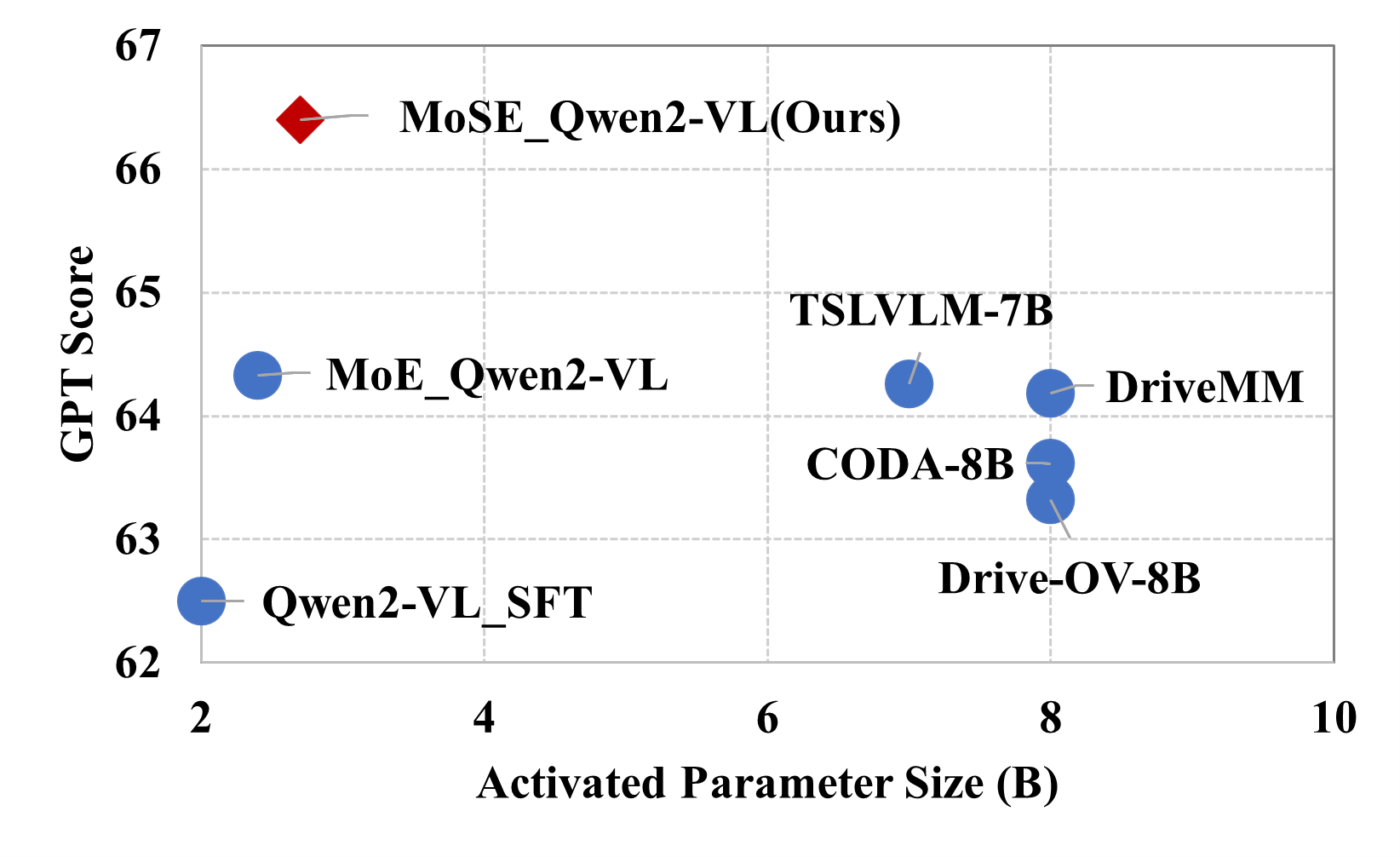}
    \caption{Activated model size and performance of various open-source specialist models on the CODA dataset. The proposed MoSE achieves SOTA performance with less than 3B parameters.
}\label{fig:modelsize}
\vspace{-0.5cm}
\end{figure}

\begin{figure*}
    \centering
    \includegraphics[width=.95\linewidth]{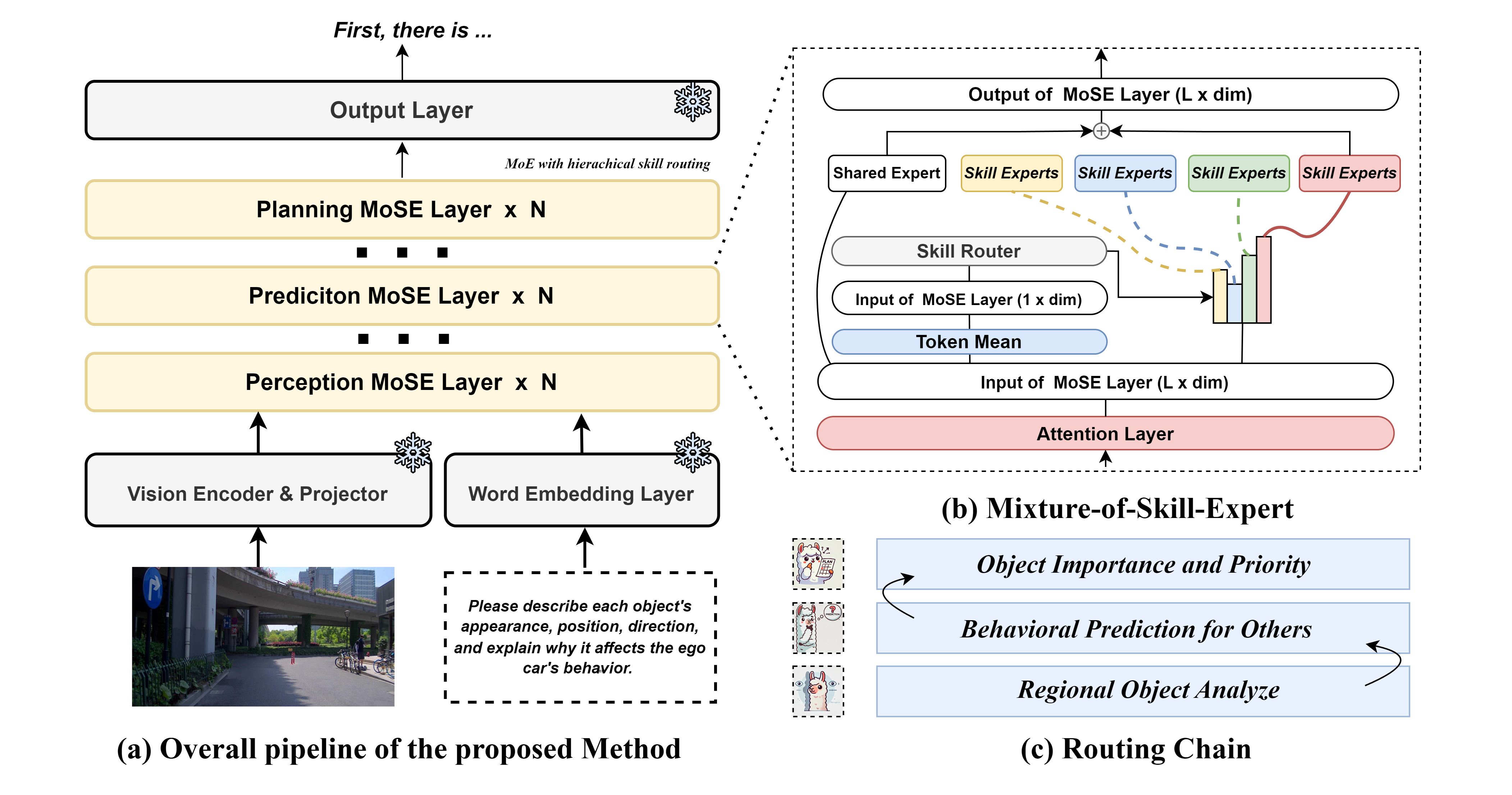}
    \caption{Overall model pipeline: (a) the step-by-step process of the model, where it progresses from perception to prediction and then to planning based on the input image and question. (b) the internal structure of each MoSE layer, where skill routing enables skill-by-skill learning. For each task, the model ultimately constructs a skill chain, which partially reflects its reasoning process, as shown in (c). For more information about the skill chain, please refer to the supplementary. In the figure, we use an autonomous driving scenario as an example; the robot follows the same process but with a different set of skill definitions.}
    \label{fig:overallpipeline}
    \vspace{-0.5cm}
\end{figure*}

Systems with human-level understanding of multi-modal scenarios have become increasingly essential for highly intelligent applications. Many downstream tasks like autonomous driving and robotics, are craving Large Language Model (LLMs) or Vision Language Model (VLMs) to not only boost the overall performance, but also to improve the system's interpretability, reasoning capabilities and interactive abilities. 
However, while web-scale pre-trained LLMs and VLMs excel at general knowledge, they often lack domain-specific knowledge required for particular tasks. Domain-specific models often require substantial design effort to effectively align task-specific details with the general knowledge encoded in pre-trained LLMs in order to achieve improved performance.

As embodied AI continues to advance toward real-world deployment in domains such as autonomous driving, home robotics, industrial automation, and humanoid agents, the demand for systems with a deep, multi-modal understanding has intensified. These systems must interpret visual inputs, comprehend complex instructions, and generate reliable, explainable actions in diverse and dynamic environments. Vision-Language Models (VLMs) and Vision-Language-Action (VLA) models have emerged as critical foundations to meet these requirements.

In embodied AI, autonomous driving and robot manipulation are representative tasks showing significant progress. For driving, LMDrive~\cite{shao2024lmdrive} improves trajectory prediction via instruction-aligned simulation data, while OmniDrive~\cite{wang2024omnidrive} and DriveLM~\cite{sima2023drivelm} leverage multi-step QA and graph-based reasoning across perception, prediction, and planning. CODA~\cite{chen2024automated} further utilizes hierarchical annotations and VLM prompting to analyze complex real-world corner cases.

In manipulation, models such as OpenVLA~\cite{kim2024openvla}, CogACT~\cite{li2024cogact}, TraceVLA~\cite{zheng2024tracevla}, and Pi0~\cite{black2410pi0} demonstrate strong generalization and zero-shot abilities by combining VLMs with action modules or vision-language-action learning frameworks.

Despite this progress, both fields share key challenges: the need for high reasoning capacity, low-latency inference, and cost-efficient deployment. Large pre-trained models remain too resource-intensive, motivating more efficient and adaptive architectures tailored to domain-specific constraints.

To address this, an emerging technique for pre-trained LLM/VLMs is the use of sparse Mixture-of-Experts (MoE) layers in Transformer-based architectures~\cite{shazeer2017outrageously}. These layers improve model efficiency and scalability by dynamically activating a small subset of specialized sub-networks, or "experts," for each input. This allows models to scale up capacity while maintaining low computational cost~\cite{lepikhin2020gshard,fedus2022switch,jiang2024mixtralexperts}. A router network determines which experts to activate based on the input, enabling flexible handling of diverse tasks and domains while utilizing only a fraction of the model's total parameters. MoE-LLaVA~\cite{lin2024moe} demonstrates that small-scale VLMs (\textless3B) equipped with MoE layers can achieve performance comparable to larger 7B models on general vision-language tasks. These insights motivate us to explore MoE-based solutions for downstream domains such as autonomous driving and robotic manipulation, where collecting large-scale, diverse training data is costly. However, naive applications of MoE in these settings often yield suboptimal results due to the limited availability of task-specific data.


In this paper, we explore the MoE technique in task-specific, small-scale ($<$3B), multi-modal vision-language models (VLMs) then introduce an efficient framework designed for different embodied AI tasks (c.f.\ Fig.~\ref{fig:overallpipeline}). Inspired by human learning strategy, we design a skill-by-skill learning approach for the routing procedure with a skill centric routing strategy.
The routing strategy identifies essential driving skills required for various driving scenarios and driving stages, guiding experts within each MoE layer to specialize in distinct scenarios and progressively develop the skills needed to address targeted conditions effectively.

Furthermore, inspired by progressive human behavior and prior end-to-end frameworks such as UniAD~\cite{hu2023planning} for autonomous driving, we further employ a hierarchical routing strategy across layers to enable the system to think and operate in a step-by-step manner. This approach not only improves consistency in question-answering across varying levels but also aligns with valuable auxiliary tasks (e.g. perception, prediction and planning for autonomous driving, and perception, interpretation and reasoning, planning for robot manipulation) in a single forward pass, eliminating the need for additional context or multi-round QA.

Experimental results demonstrate that, with only 2B+ parameter size (e.g.\ Qwen2VL-2B~\cite{wang2024qwen2}), our model achieves performance competitive to other state-of-the-art autonomous driving reasoning tasks and robot manopulation tasks with more than 8B+parameters. Without any acceleration techniques, our model improves inference speed by at least 15.7\%.
To sum up, our contribution can be summarized as follows: \par
•	We propose a novel Mixture-of-Skill-Experts (MoSE) to enhance the reasoning capability of small-scale VLMs ($<$3B parameters) for embodied AI tasks like autonomous driving and robot manipulation. Inspired by the learning process of human drivers, our approach provides more interpretable reasoning capabilities. 
~\par
•	We propose a skill-oriented routing mechanism, enabling the model to reason step-by-step and guiding experts to learn scenario-specific tasks skill-by-skill. The skill-based architecture enhances adaptability and safety, especially in dynamic factory and service environments, where tasks often involve clearly structured skill requirements.
~\par
•	Compared to existing methods based on open-source models and data, MoSE achieves state-of-the-art performance on AD corner case reasoning task and robot manipulation reasoning task with significantly reduced activated parameter size (at least by $62.5\%$). It offers a highly efficient AI solution for real-world applications.

\section{Related Work}

\paragraph{LLM/VLM for Embodied Autonomous Machines}
Large Language Models (LLMs) and Vision-Language Models (VLMs) have become core components in embodied AI, spanning domains like autonomous driving and robotic manipulation. In driving, early methods focus on scene understanding using closed-source models such as GPT-3.5 and GPT-4V~\cite{mao2023gpt,sha2023languagempc,wen2023road}. LLM-Driver~\cite{chen2024driving} treats LLMs as multi-task learners predicting both descriptions and control commands. With the rapid development of multi-modal models like LLaVA~\cite{liu2023llava,liu2023improvedllava,liu2024llavanext}, recent efforts leverage large-scale annotated datasets~\cite{shao2024lmdrive,sima2023drivelm,wang2024omnidrive}. LMDrive and OmniDrive align features from tailored visual encoders~\cite{zhai2023sigmoid,li2023blip} to LLMs, while DriveLM~\cite{sima2023drivelm} builds a Graph VQA dataset on nuScenes~\cite{nuscenes2019} and Carla for reasoning and trajectory prediction. DriveMoE~\cite{yang2025drivemoe} further introduces MoE-based routing for multi-camera input and behavior-specific modules.

In robotic manipulation, the focus is shifting from modular pipelines to generalist end-to-end policies. OpenVLA~\cite{kim2024openvla} directly fine-tunes VLMs to predict actions from vision-language input. Later models such as CogACT~\cite{li2024cogact} and TraceVLA~\cite{zheng2024tracevla} integrate diffusion-based policy modules and visual trace prompting. Pi0~\cite{black2410pi0} replaces end-to-end decoding with a separate action expert trained via flow matching. While these approaches improve generalization, they still rely on domain-specific fine-tuning for complex tasks. Altogether, LLM/VLM-driven frameworks are enabling more scalable embodied AI, though challenges remain in adapting to diverse environments and tasks efficiently.


\paragraph{Multimodal Mixture-of-Experts}
Mixture-of-experts architectures with sparse activations of experts provide a possible way to scale up LLMs with high efficient training and inference. In the transformer based architectures, one promising way is to substitute feed-forward network (FFN) layers with sparse MoE layers~\cite{shazeer2017outrageously,zoph2202st,fedus2022switch}.

MoE-LLaVA~\cite{lin2024moe} builds light weight VLMs using MoE and gain comparable performance on various visual understanding  tasks with fewer parameters.
\cite{Chen_2023_ICCV} automatically determines the number of activated experts for each task, avoiding the laborious manual tuning of optimal model size. 
\cite{yun2024flex} proposes to flexibly incorporate arbitrary modality combinations while maintaining robustness to missing data. The router specializes in handling fewer modality combinations by assigning the Top-1 gate to the expert corresponding to the observed modality combination.
Med-MoE~\cite{jiang2024moe} introduces a lightweight framework for multi-modal medical tasks, addressing both discriminative and generative needs in the medical domain. They proposed to pre-train the router and apply domain-specific expert fine-tuning. 
Modality-aware MoE~\cite{lin2024moma} divides experts into modality-specific groups (one group for text, one for vision). Each group can focus on the features of its modality while still allowing cross-modal interactions in shared layers. 

\begin{figure}
    \centering
    \includegraphics[width=0.95\linewidth]{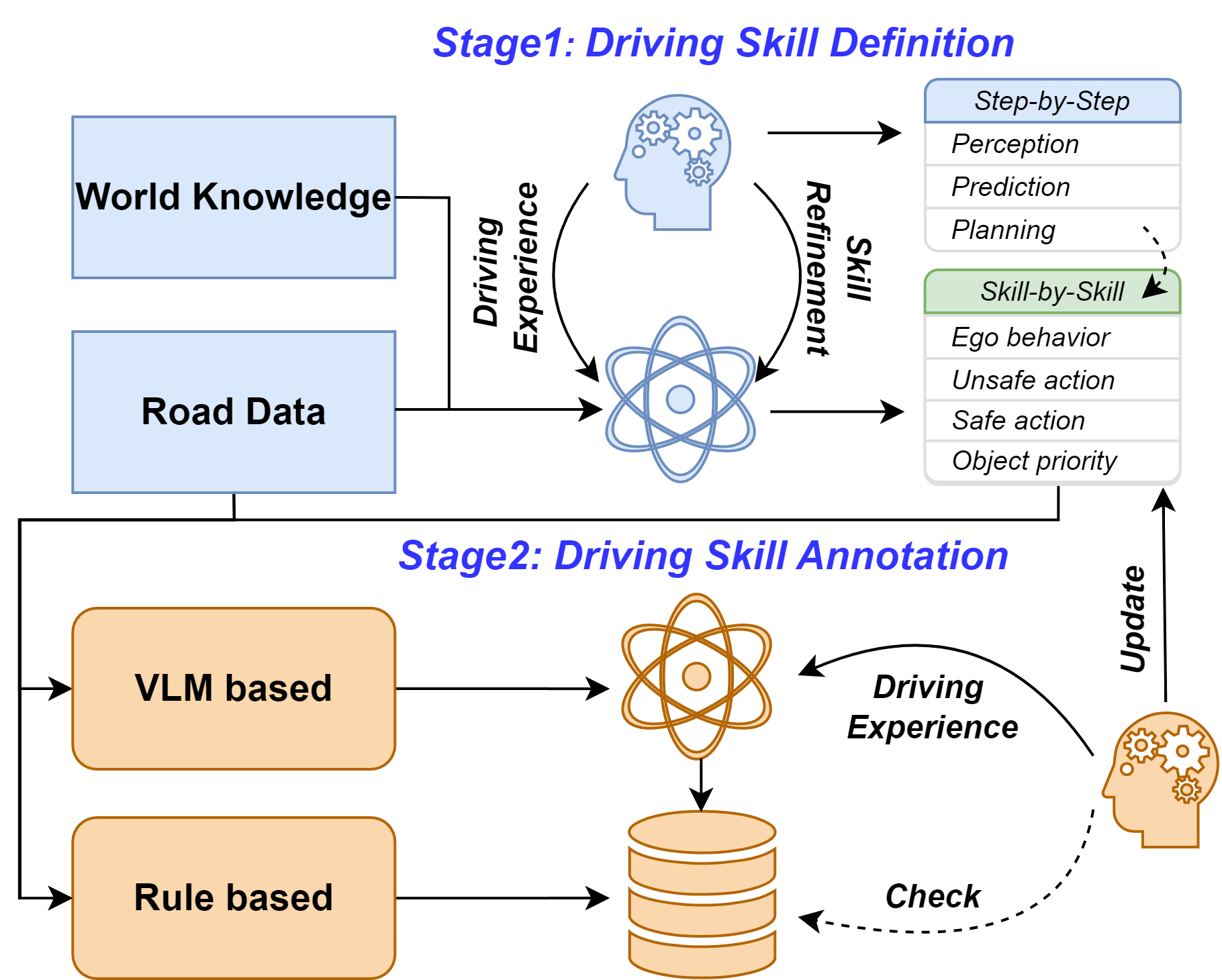}
    \caption{The process of constructing skill data consists of two stages. In the first stage, we define the skills relevant to the current task, specifically autonomous driving. In the second stage, we perform data annotation using both rule-based and VLM-based methods.
    }
    \vspace{-0.5cm}
    \label{fig:Figure-4}
\end{figure}

\section{Method}
In this section, we present the proposed Mixture-of-Skill-Experts method. Given a visual signal $I_v$ and a user instruction $I_t$, we develop a MoSE based VLM $F(\cdot  )$ that generates driving reasoning result:
\begin{equation}
R_t = F(I_v,I_t). \label{0}
\end{equation}
The input visual signal $I_v$ can be single-view or multi-view images or videos, while $I_t$ encompasses mostly questions about embodied AI perception, prediction, planning, reasoning and suggestion.

As illustrated in Fig.~\ref{fig:overallpipeline}(a), the overall system processes multimodal inputs, comprising visual observations in image format and language-based interactive inputs in text format, and generates a text-based output.
We adapt general MoE into skill-by-skill learning at both the case and task levels, each task is decomposed into hierarchical levels, and for each case, different skills are activated at different levels. We begin with language-visual-aligned, instruction-tuned, small-scale VLMs, inserting MoE layers, then initiating the router pre-training with a small amount of labeled data.
This step enables the router to focus on skill routing rather than token-level distinctions. The router learns to characterize necessary skills in each stage and activate appropriate skill experts. After the router training, we conduct supervised fine-tuning on the entire model. 
Following the approach of Qwen2-VL, we employ a pre-trained Vision Transformer~\cite{dosovitskiy2020image} (ViT) as the vision encoder, maintaining the existing processing pipeline for vision tokens. 
During training, we only finetune the MoSE blocks while keeping the rest of the model frozen.  

\subsection{Skill Centric Routing Mechanism}

In a general MoE, a router network learns to select activated experts for each individual token automatically. The trained router tends to focus on distinguishing different domains or modalities which may be inadequate for some tasks like autonomous driving and robot manipulation. Compared to internet-scale data, specific embodied AI images and videos exhibit higher similarity, with questions following specific patterns. These factors introduce additional challenges for MoE training, necessitating a carefully designed architecture and learning strategy to address limitations in data scale and diversity.

To enable skill-by-skill learning, the router must gain a refined understanding of both the driving scene and input text, facilitating expert selection for each necessary stage based on the overall driving context. To mimic human reasoning and learning process, we design a skill centric routing mechanism to guide the model to learn and think skill-by-skill. To build such a procedure, we start by defining essential skills for our task, autonomous driving. One thing need to be mentioned is that the skills can be shared across different datasets. After having the skills, we perform data annotation using rule based method or other large VLMs, in our case, GPT-4o. The data quality is directly related to skill definition, so one can optimize the human annotation effort by adjusting the skills. After having the skill data, we pretrain the router of all target layesr, the trained router can guide the experts follow a hierarchical path and generate answer step-by-step.

\subsubsection{Skill Definition}\label{sec:skilldef}
To enable a skill-centric routing mechanism, we begin by defining driving skills based on two principles:
\begin{enumerate}
    \item Minimize the complexity introduced by the router to simplify routing.
    \item Ensure skills provide complete data coverage and generalize across multiple datasets.
\end{enumerate}

To meet these goals, we introduce a general skill at each hierarchical level. This ensures coverage for ambiguous samples and handles cases where certain levels are unused by assigning general skills to them.

As shown in Fig.~\ref{fig:Figure-4} Stage 1, we use state-of-the-art language models (e.g., GPT) to define driving and manipulation skills. GPT receives examples from diverse datasets and is supplemented with human driving and manipulation knowledge. The model generates skill definitions at multiple levels, which are further refined by human experts.

Our model activates appropriate skills at each hierarchical level, forming a structured skill chain (Fig.~\ref{fig:overallpipeline}(c)). This chain supports both MoSE’s reasoning and training, and improves interpretability during inference. For instance, the model first detects key objects, predicts their possible behaviors, and evaluates their relevance. This structured reasoning helps identify influential objects and provides interpretability for debugging and user trust.

\subsubsection{Skill Data Annotation}\label{sec:skillanno}
Based on the defined skills, we sample a subset of data from the target dataset and annotate each instance with its corresponding skill. Depending on the data characteristics, we apply either rule-based or VLM-based (e.g., GPT) annotation methods, as shown in Fig.~\ref{fig:Figure-4} Stage 2.

For structured tasks and general scenarios, we adopt rule-based annotation. These rules map data to specific skills—for instance, distinguishing whether a question targets the full image or a specific region. This approach is cost-effective, yields high accuracy, and supports large-scale annotation with minimal manual effort.

In cases where rules are insufficient, we use GPT-based annotation. The input includes the scene image, ego vehicle state, and ground-truth trajectory. A predefined rule list is included in the prompt, specifying criteria such as assigning the general skill to incomplete samples.

We iteratively refine this rule list, both manually and automatically, to improve annotation quality. This results in high-quality labeled data with minimal manual intervention. Annotation prompts are provided in the supplementary material. Although skill annotation introduces overhead, we demonstrate that MoSE requires only a small subset of annotated data to learn the hierarchical routing mechanism. Moreover, for GPT-based annotations, manual verification is often unnecessary, as annotation challenges are already addressed during the skill definition stage.

\begin{figure}
    \centering
    \includegraphics[width=0.95\linewidth]{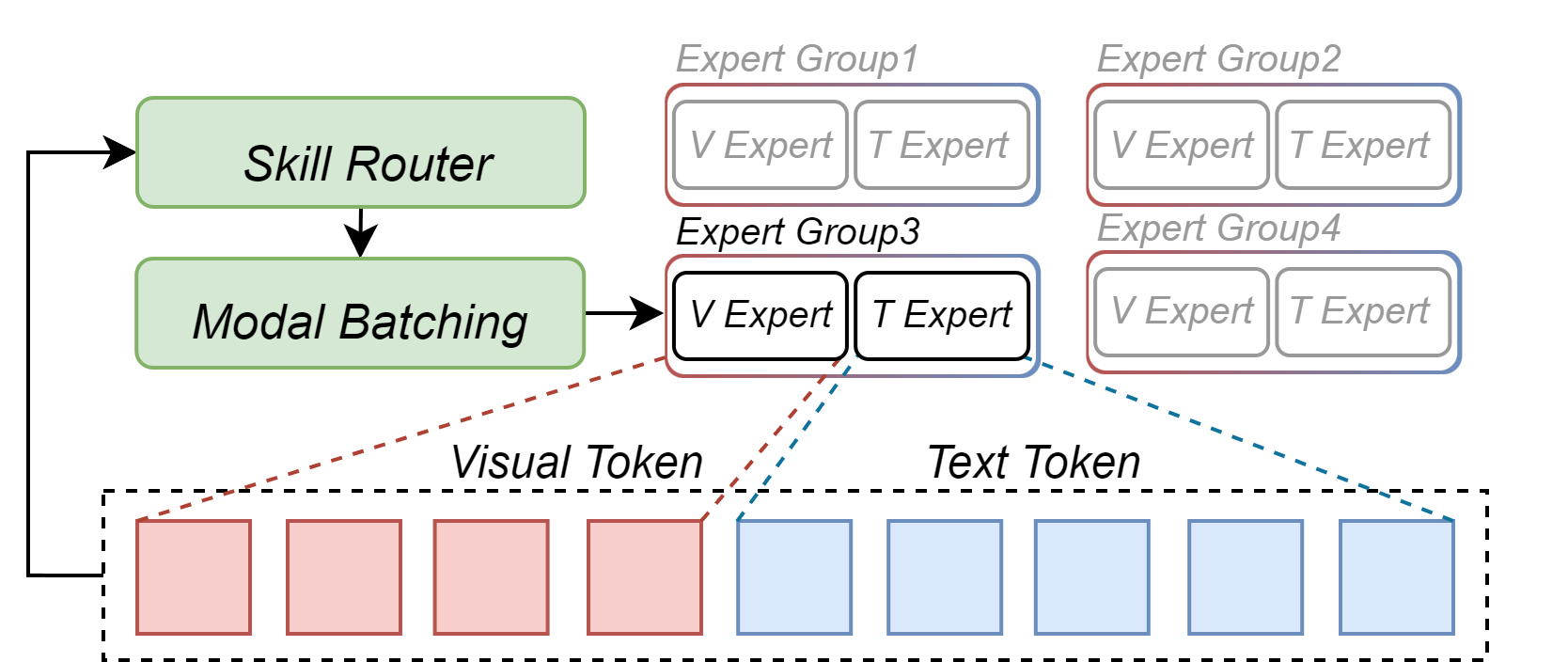}
    \caption{Pipeline for multi-modal scaling up. This approach enhances the model's capability without requiring additional data annotation or pretraining.}
    \vspace{-0.5cm}
    \label{fig:scale up}
\end{figure}


\begin{figure*}
    \centering
    \includegraphics[width=0.95\linewidth]{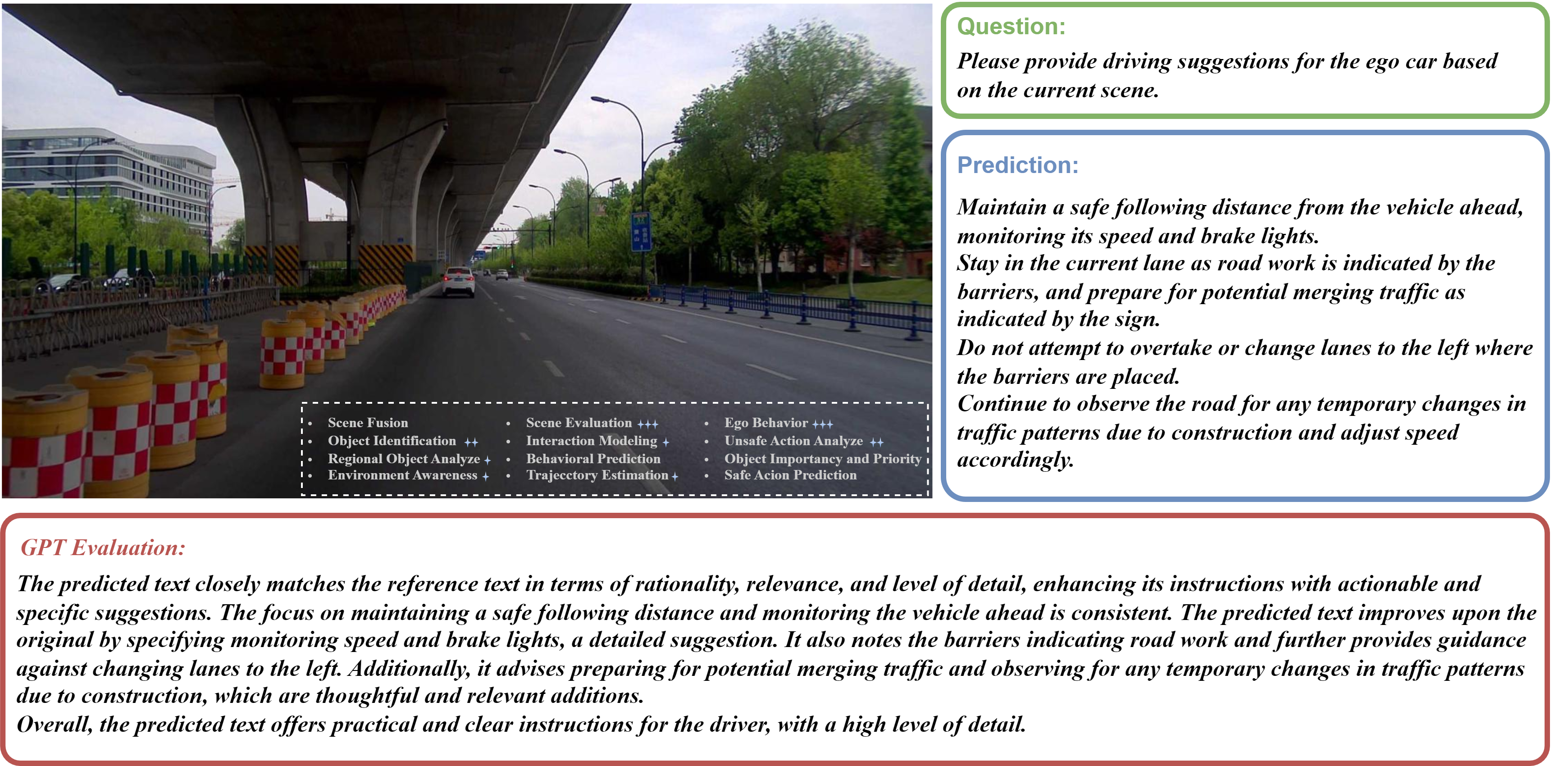}
    \caption{Visualization for reasoning results, where GPT Evaluation reflects the GPT model's assessment of the generated results. The white-boxed region in the image shows the expert skills activated at different MoSE layers.}
    \vspace{-0.3cm}
    \label{fig:visualization}
\end{figure*}

\subsubsection{Router Pretraining}\label{sec:skillpretrain}
After having the skill annotation data, inspired by recent work~\cite{jiang2024moe} 
we pretrain the routers in different hierarchical layers. To further enhance the efficiency, we take the mean token of all tokens from the previous layer  $T_{Mean}$. Taking $T_{Mean}$ as input, the training loss function can be expressed as Eq.~\eqref{1}. 
\begin{equation}
L_{Router} = -\sum_{i=1}^{n} y_i\log p(y_i|T_{Mean}). \label{1}
\end{equation}

\subsection{Expert Networks and Training}\label{sec:MoSE}
After constructing the hierarchical skill routing mechanism, 
we construct the MoSE layers with one shared expert and multiple skill-specialized experts.
The shared expert is consistently activated to capture global information, enhancing the model's robustness. The proposed skill centric routing mechanism ensures that the experts not only develop scenario-specific skills but also acquire distinct skills for different stages of reasoning. 

Following standard practices, we adopt a sparse MoE configuration where the original FFN layers in all even-numbered Transformer layers were replaced with the proposed MoSE modules, as can be seen in Fig. ~\ref{fig:overallpipeline}(b). This design balances model scalability with the total number of parameters and active parameters.


We optimize the output of LLM through a generative loss in an auto-regressive manner. Given the input image and text, the model progressively generates each element $Y$. For all trainable parameters $\theta$, the loss can be expressed as:
\begin{equation}
 L_{reg} = -\sum_{i=1}^{N}\log\,p_{\theta } (Y^{[P+i]}|V,T^{[:i-1]}),  \label{2}
\end{equation}
where $Y$ is the output sequence, $K=P+N$ is the total output length. We only calculate the loss for $N$ elements which is the newly generated text.

\paragraph{Scaling-Up on MoSE}
To further enhance the model's capabilities and scale it up without retraining the router, we propose 
MoSE with expert groups.
The overall model pipeline is illustrated in Fig.~\ref{fig:scale up}.

Based on the routing results from the pretrained router, we construct a group of experts, where multiple experts $n$ are assigned to $n$ sub-skills. Specifically, for the current task, we adopt a modality-specific expert strategy with $n=2$ where different experts handle different modalities e.g.\ text or image.
That's to say, we assign two experts within each expert group: one for processing image tokens and the other for processing text tokens under the corresponding specific skill. This strategy only involves skill experts,  with the shared experts remaining unchanged.
\section{Experimental Results}

\paragraph{Dataset Settings}
We conduct our autonomous driving experiments on the CODA dataset~\cite{chen2024automated} which is designed for autonomous driving reasoning task. The dataset collects driving scenarios and extract objects which may influence ego vehicle, the final driving suggestion is generated considering all key objects. Compared to other driving reasoning dataset, CODA focus on multi-modal corner cases and consider the hierarchy data structure. It requires models to generate relatively long answers containing description and analyzing of multiple objects. In this paper, we follow the official training and testing settings, and use GPT-4o for evaluation. 

We conduct our robot manipulation reasoning experiments on Robo2VLM\cite{chen2025robo2vlm} dataset which uses rich, real and multi-modal robot trajectory data to enhance and evaluate VLMS, especially reasoning ability. The dataset cover 463 manipulation scenes and 3396 robot manipulation tasks, which can benchmark and improve VLM capabilities in spatial and interaction reasoning.

\begin{figure}[tb]
    \vspace{-0.3cm}
    \centering
    \includegraphics[width=0.9\linewidth]{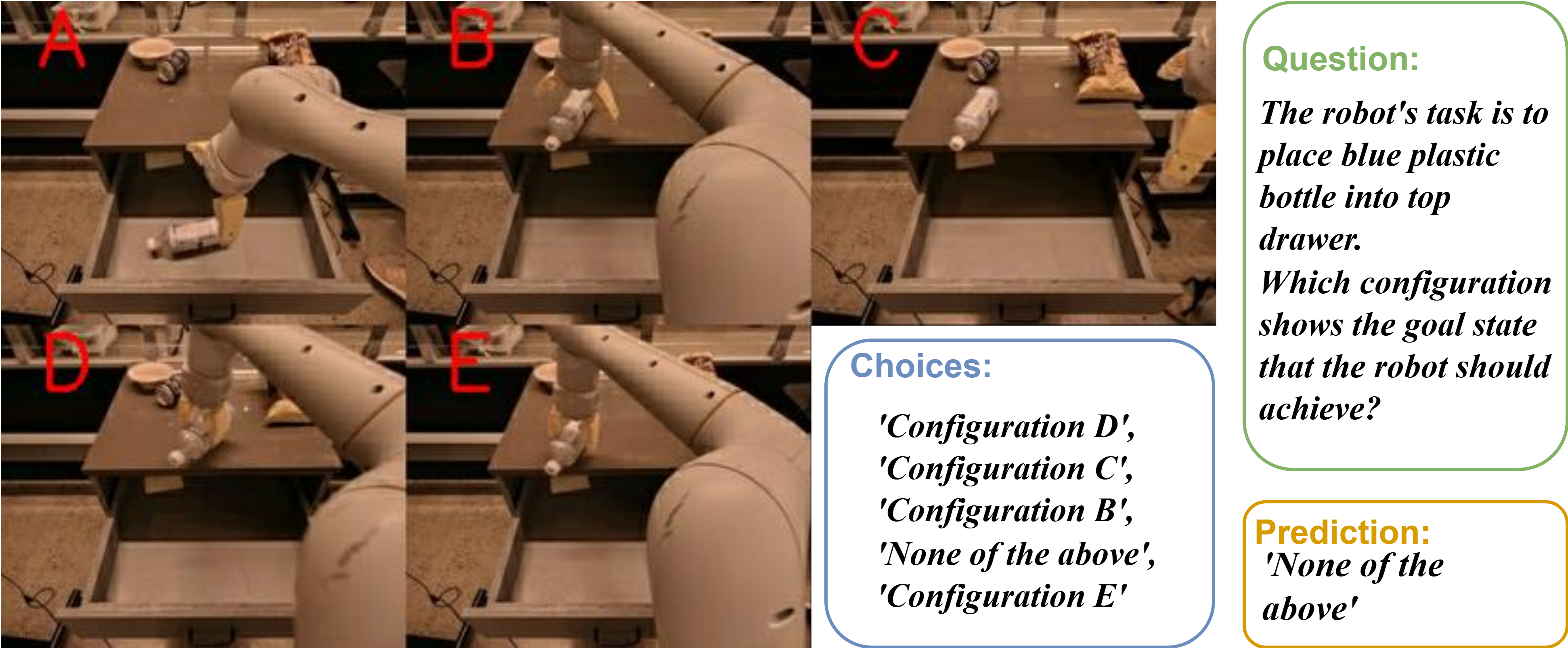}
    \caption{Visualization of the Robot Manipulation reasoning. For each robotic scenario, the model identify the correct answer from a set of five candidates, which include distractor options designed to test its reasoning capabilities.}
    \vspace{-0.5cm}
    \label{fig:visualization}
\end{figure}

\paragraph{Model Settings}
We conduct experiments on both MoSE tasks based on Qwen2-VL-2B~\cite{bellagente2024stable} which is a small-scale VLM. 
We adapt a total of $14$ MoSE layers. Experimental results indicate that a relatively uniform hierarchical distribution yields better performance. Therefore, in our final configuration, we allocate $[4,5,5]$ layers to perception, prediction, and planning for AD, and perception, interpretation and reasoning, and planning for robot tasks, respectively.

The input image size is set to $[800,450]$ to balance the image quality and GPU memory. We follow the sparse setting of general MoE that only half of the layers are equipped with MoSE. During the pretraining stage and finetuning stage, we set expert number to $4$ and use $1$ shared expert. compare to Qwen-vl-2B base model, it brings 400M extra activate parameters for MoSE and 800M for MoSE with expert group. Additionally, we also experiment with general MoE method on same backbone model (denoted as Qwen2-VL\_MoE in the result table), with same number of experts and number of activated experts to our MoSE. To enable a better comparison, we take CODA as an example and finetune a QwenVL2-2B under a supervised finetuning settings (denoted as Qwen2-VL\_SFT). For hierarchical skill routing, we annotate $2000$ skill samples for both AD and robot using rule-based methods to pretrain the routers.

\subsection{Experiment results on CODA}

\begin{table}[t]
\centering
\renewcommand{\arraystretch}{1.2}
\resizebox{\linewidth}{!}{
\begin{tabular}{lcccc}
\toprule
\textbf{Model} & \textbf{General} & \textbf{Regional} & \textbf{Suggest.} & \textbf{Score} \\
\midrule
\multicolumn{5}{c}{\textit{Closed-Source Models}} \\
\midrule
\rowcolor{gray!10} Gemini-Pro~\cite{gemini2024} & 25.24 & 51.38 & 27.40 & 34.67 \\
\rowcolor{gray!10} GPT-4V~\cite{openai2023gpt4v} & 57.50 & 56.26 & 63.30 & 59.02 \\
\rowcolor{gray!10} TSLVLM+GPT4V~\cite{xue2024two} & 58.70 & 83.41 & 74.26 & 72.12 \\
\rowcolor{gray!10} RtoG(GPT-4o)~\cite{han2024regional} & 59.00 & 84.37 & 70.80 & 71.39 \\
\midrule
\multicolumn{5}{c}{\textit{Generalist Models}} \\
\midrule
Shikra-7B~\cite{chen2023shikra} & 12.24 & 22.94 & 10.20 & 15.12 \\
LLaVA1.5-7B~\cite{liu2023improvedllava} & 19.30 & 42.06 & 23.16 & 28.17 \\
Qwen-VL-Chat-7B~\cite{bai2023qwen} & 18.22 & 26.62 & 22.06 & 22.30 \\
MiniCPM-V-2.5-8B~\cite{yao2024minicpmv} & 41.12 & 57.20 & 48.48 & 48.93 \\
LLaVA1.5-13B~\cite{liu2023improvedllava} & 24.54 & 42.41 & 27.90 & 31.61 \\
LLaVA-NeXT-13B~\cite{liu2024llavanext} & 29.86 & 53.63 & 31.92 & 38.47 \\
InternVL-V1-5-20B~\cite{chen2024internvl} & 38.38 & 61.53 & 41.18 & 47.03 \\
Qwen-VL-Max~\cite{qwen-vl-max} & 34.60 & 68.17 & 67.40 & 56.72 \\
LLaVA-one~\cite{li2024llavaonevision} & 38.70 & 51.70 & 49.32 & 46.57 \\
\midrule
\multicolumn{5}{c}{\textit{Specialist Models}} \\
\midrule
CODA-8B~\cite{chen2024automated} & 55.04 & 77.68 & 58.14 & 63.62 \\
Mini-Driver-83M~\cite{zhang2024minidrive} & 21.60 & 62.15 & 45.40 & 43.05 \\
Mini-Drive-137M~\cite{zhang2024minidrive} & 24.60 & 66.34 & 45.44 & 45.46 \\
TSLVLM-7B~\cite{xue2024two} & 52.84 & \textbf{78.33} & 61.62 & 64.26 \\
DriveMM-8B~\cite{huang2024drivemm} & 52.94 & 77.76 & 61.84 & 64.18 \\
Drive-OV-8B~\cite{huang2024drivemm} & 53.60 & 77.64 & 58.72 & 63.32 \\
\rowcolor{gray!15} Qwen2-VL\_SFT & 52.70 & 69.50 & 65.30 & 62.50 \\
\rowcolor{gray!15} Qwen2-VL\_MoE\_41 & 55.00 & 71.10 & 66.90 & 64.33 \\
\rowcolor{gray!15} Qwen2-VL\_MoE\_81 & 47.20 & 66.80 & 59.20 & 57.73 \\
\rowcolor{yellow!20} MoSE (Ours) & \textbf{58.10} & 71.10 & 68.90 & 66.03 \\
\rowcolor{yellow!20} MoSE w. Exp Grp (Ours) & 56.40 & 72.40 & \textbf{70.30} & \textbf{66.40} \\
\bottomrule
\end{tabular}}
\caption{Performance comparison of different models.yellow rows are our models. Bold denotes best in group. W.Exp Grp is the scaling-up MoSE model}
\label{tab:model_performance}
    \vspace{-0.3cm}
\end{table}

Table~\ref{tab:model_performance} shows our results on CODA compared to other state-of-the-art models. CODA contains different subsets, general perception task lies in a comprehensive understanding of critical road key entities in driving scenarios (denoted ``General''), regional perception tasks focus on understanding corner case objects when provided with specific bounding boxes (denoted ``Regional''), and driving suggestions task aims to formulate driving advice (denoted ``Suggest.''). The results are evaluated using GPT-4o, which is the same evaluation model according to CODA (denoted as ``Score''). The reported results contain generalist open-source and closed-source models, as well as specialist models which are supervised finetuned on the CODA training set. As can be seen from the table, with less than 3B parameters, the proposed MoSE outperform other SOTA methods with far fewer parameters. TSLVLM$+$GPT4V has the best performance but it relies on GPT for refinement.

At the bottom of the table, we compare MoSE with baseline methods, MoE, and SFT. The results indicate that incorporating MoE for scaling up the base model effectively improves performance. However, it also increases training difficulty and imposes higher demands on data.

From the results of MoE with 8 experts, 
the performance decline suggests that the CODA training set is insufficient to fully support the model's training, which empirically demonstrates that general straightforward application of MoE encounters challenges and often results in suboptimal performance. In contrast, the proposed MoSE outperforms MoE. 
Further, the MoSE model with expert groups, despite having the same training data as MoE-8-1, does not suffer from performance degradation. Instead, it achieves a slight improvement, which we attribute to effective data utilization brought by hierarchical skill routing.

Beyond the performance improvements demonstrated by the evaluation results on driving suggestion task, One thing need to be mentioned is that, the official setting of CODA uses general result and object result of context to generate the final driving suggestion. In our settings, we conduct single-round QA without multi-turn conversation and long context.
\begin{table}[t]
\centering
\renewcommand{\arraystretch}{1.2}
\footnotesize  
\label{tab:robo2vlm_performance}

\resizebox{0.7\linewidth}{!}{
\begin{tabular}{l c}
\toprule
\textbf{Model} & \textbf{Score} \\
\midrule
\multicolumn{2}{c}{\textit{Generalist Models}} \\
\midrule
LLaVA1.5-7B~\cite{liu2023improvedllava}          & 21.58 \\
LLaVA1.6 Mistral-7B~\cite{liu2023improvedllava} & 24.09 \\
LLaVA1.6-34B~\cite{liu2023improvedllava}        & 24.94 \\
Llama 3.2-90B~\cite{grattafiori2024llama3}       & 28.60 \\
Qwen-2.5VL-7B~\cite{qwen-vl-max}                & 30.63 \\
Qwen-2.5VL-13B~\cite{qwen-vl-max}               & 37.68 \\
Qwen-2.5VL-72B~\cite{qwen-vl-max}               & 37.76 \\
\midrule
\multicolumn{2}{c}{\textit{Specialist Models}} \\
\midrule
\rowcolor{gray!10} MoE                         & 33.57 \\
\rowcolor{yellow!20} \textbf{MoSE (Ours)}     & \textbf{47.65} \\
\bottomrule
\end{tabular}}
\caption{Performance on Robo2VLM. MoSE achieves significant performance improvement compared to MoE.}
\vspace{-0.2cm}
\end{table}

\begin{table}
\centering
\renewcommand{\arraystretch}{1.2}
\resizebox{.9\linewidth}{!}{
\begin{tabular}{c|c|c|c|c}
\hline
Skill Data Size & General & Regional & Suggest. & Score \\ 
\hline
2000  & 58.1 & 68.9 & 71.1 & 66.03 \\ \hline
3000   & 56.7 & 68.9 & 70.7 & 65.43 \\ \hline

\hline
\end{tabular}}
\caption{Comparison on different skill-annotated data size. Additional skill data does not result in a performance gain.}
\label{tab:skill-data-size}
    \vspace{-0.5cm}
\end{table}

\begin{figure*}
    \centering
    \includegraphics[width=0.24\linewidth,trim=.5cm .0cm .0cm .2cm, clip]{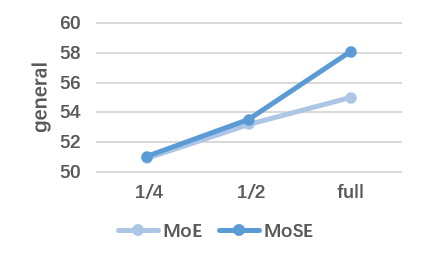}
    \includegraphics[width=0.24\linewidth,trim=.5cm .0cm .0cm .2cm, clip]{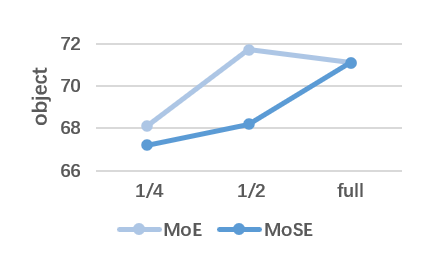}
    \includegraphics[width=0.24\linewidth,trim=.5cm .0cm .0cm .2cm, clip]{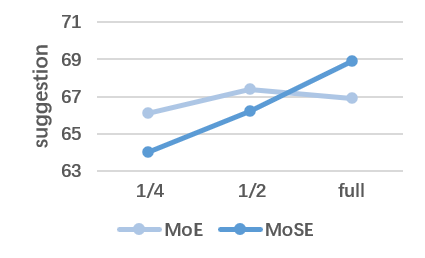}
    \includegraphics[width=0.24\linewidth,trim=.5cm .0cm .0cm .2cm, clip]{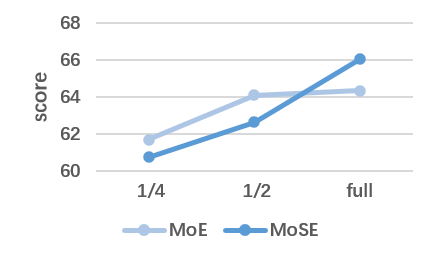}
    \caption{Performance variations of MoE and MoSE after training on different data scales. MoSE demonstrates better performance scaling than MoE as data size increases, highlighting its potential for larger and more complex datasets.}
        \label{fig:data size}
    \vspace{-0.3cm}
\end{figure*}

 \subsection{Experiment results on Robo2VLM}

Table~\ref{tab:model_performance} shows our results on the Robo2VLM dataset. Robo2VLM contains multiple-choice Visual Question Answering data from human-teleoperated robot trajectorys. The score in Robo2VLM refers to the accuracy on the testing set, where each question is a 5-way multiple-choice and only one option is correct. Robo2VLM contains varying numbers of input images and covers a large diversity of scenes, which poses greater challenges for relatively small models. Compared to the suboptimal performance of the MoE baseline, our proposed skill-centric approach achieves more than 40\% improvement in performance.

\begin{figure}
    \centering
    \includegraphics[width=0.49\linewidth]{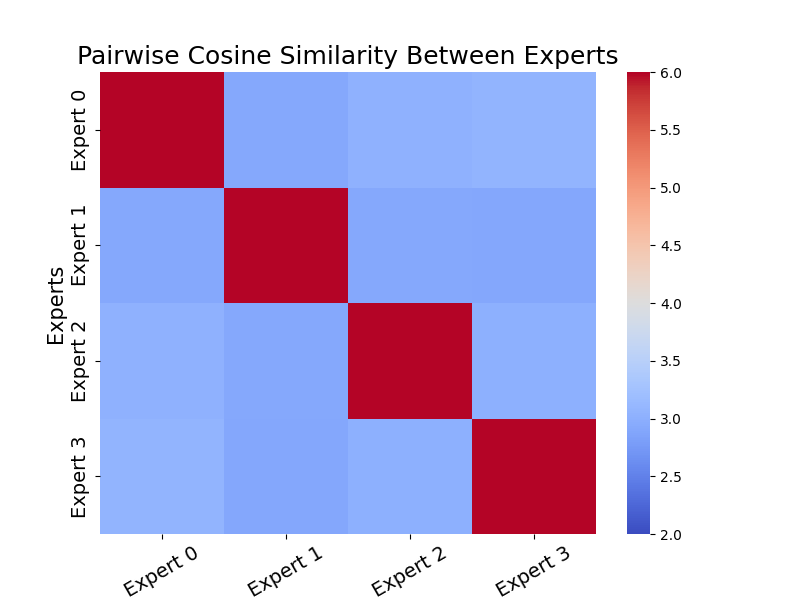}
    \includegraphics[width=0.49\linewidth]{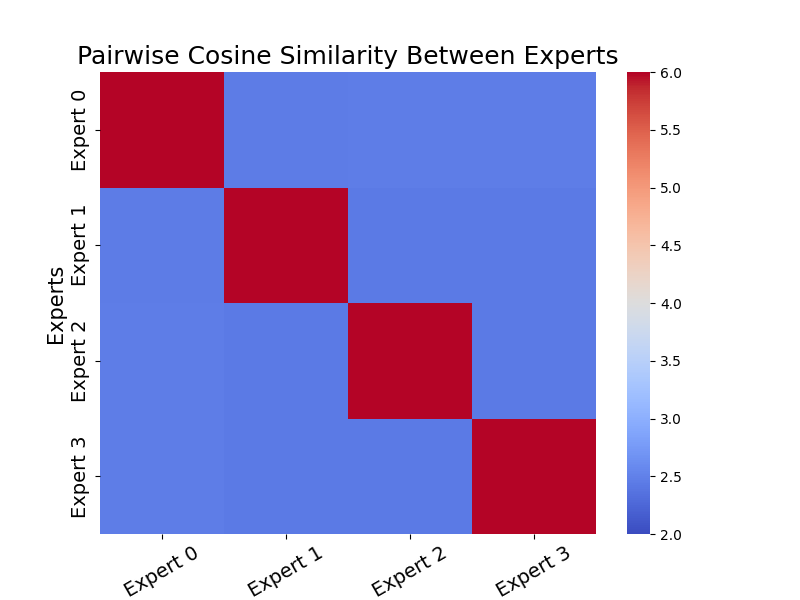}
    \caption{Visualization of expert weight distribution similarity in the last layer. Left:Moe, Right:MoSE. The expert distribution in the MoSE model is more diverse. The similarity values are log-transformed for better clarity in reflecting differences. Visualizations for additional layers can be found in the supplementary.}
        \label{fig:visualization}
\vspace{-0.5cm}
\end{figure}



\subsection{Further Analysis}
\paragraph{Training data size}
In Fig.~\ref{fig:data size},we discuss model performance with different training data size. During  training, we randomly sample half and quarter QAs from CODA and use the official testing set for evaluation. As the data scale increases, we observe that the performance of the MoE model peaks when trained with half of the data, with no further improvements as the data size continues to grow. We hypothesize that this plateau may be attributed to the way experts are allocated. In contrast, MoSE exhibits a steady improvement in performance, with and without scaling up. This suggests that as the dataset size increases, MoSE has the potential to achieve even better training results.

\paragraph{Data Size for Skill Annotation }
Given that the proposed MoSE, compared to the original MoE, relies on a small annotated data subset, we may wonder whether this additional requirement could limit the model’s applicability to other datasets or tasks, or whether it would hinder its scalability. In the experiments shown in Table~\ref{tab:skill-data-size}, we pretrain the model’s router using $2000$ and $3000$ skill-labeled samples, respectively. The results indicate that further scaling up the skill dataset does not lead to additional performance gains, thereby confirming that MoSE is not constrained by the need for extensive additional annotations.

\paragraph{Diversity of expertise}

We visualize the similarity of each expert in MoE layers. In particular, we analyze the weights of one MLP layer within the MoE modules. In a MoE framework, greater differentiation among experts enhances model performance by promoting specialization and efficient resource allocation. When experts are highly distinct, each one is able to focus on a specific subset of tasks or input distributions, leading to more refined representations and better decision-making.This helps mitigate redundancy, ensuring that experts contribute unique knowledge rather than overlapping in functionality. Additionally, a more diverse set of experts reduces competition for the same data, allowing the router to make more discriminative assignments, improving both training efficiency and generalization. As shown in the Fig.~\ref{fig:visualization}, with the hierarchical skill routing strategy, our experts have more diverse weights.





\section{Conclusion}

In this paper, we proposed a task oriented Mixture-of-Experts method named Mixture-of-Skill-Experts, which mimics human learning and the reasoning process and learns from a dataset step-by-step, skill-by-skill. 
Empirically, MoSE demonstrates that a thoughtfully designed, skill-oriented routing mechanism can achieve superior performance in embodied AI tasks, such as autonomous driving and robot manipulation. 
Furthermore, a scaling-up strategy is proposed to further boost the model's capacity without additional data or pretraining. 
Further experiments show that MoSE's hierarchical skill routing not only enhances performance with increasing data sizes but also maintains scalability by necessitating only a small number of additional layers within the expert groups.

\bibliography{aaai2026}

\end{document}